\documentclass[sigconf]{acmart}

\usepackage{graphicx}
\usepackage{balance}
\usepackage{subfigure}

\usepackage{amsmath}
\usepackage{mathtools}
\usepackage{amsthm}
\usepackage{multirow}
\usepackage{colortbl}
\usepackage{algorithm,algorithmic}
\usepackage[capitalize,noabbrev]{cleveref}

\usepackage{mathtools}
\usepackage{tikz}

\AtBeginDocument{%
  }

\setcopyright{acmlicensed}
\copyrightyear{2025}
\acmYear{2025}
\setcopyright{acmlicensed}\acmConference[MM '25]{Proceedings of the 33rd
ACM International Conference on Multimedia}{October 27--31, 2025}{Dublin,
Ireland}
\acmBooktitle{Proceedings of the 33rd ACM International Conference on
Multimedia (MM '25), October 27--31, 2025, Dublin, Ireland}
\acmDOI{10.1145/3746027.3755449}
\acmISBN{979-8-4007-2035-2/2025/10}
\begin{document}

\title{BSGS: Bi-stage 3D Gaussian Splatting for Camera Motion Deblurring}

\author{An Zhao}  
\orcid{0009-0007-1183-3402}

\affiliation{
 \institution{ Nanjing University of Aeronautics and Astronautics} 
 \city{Nanjing}
\country{China}
 }
  \email{anzhao@nuaa.edu.cn}

\author{Piaopiao Yu} 
\authornotemark[1]
\orcid{0000-0003-2093-7792}

\affiliation{
 \institution{ Nanjing University of Aeronautics and Astronautics} 
  \city{Nanjing}
\country{China}
 }
  \email{yupiaopiao@nuaa.edu.cn}
 
 \author{Zhe Zhu}  
 \orcid{0000-0001-5314-9392}

\affiliation{
 \institution{ Nanjing University of Aeronautics and Astronautics}
  \city{Nanjing}
\country{China}
 }
  \email{zhuzhe0619@nuaa.edu.cn}
 
 \author{Mingqiang Wei} 
 \authornote{Corresponding Author} 
 \orcid{0000-0003-0429-490X}

\affiliation{
 \institution{ Nanjing University of Aeronautics and Astronautics}
  \city{Nanjing}
\country{China}
 }
  \email{mqwei@nuaa.edu.cn}

\begin{abstract}

3D Gaussian Splatting has exhibited remarkable capabilities in 3D scene reconstruction.
However, reconstructing high-quality 3D scenes from motion-blurred images caused by camera motion poses a significant challenge.

The performance of existing 3DGS-based deblurring methods are limited due to their inherent mechanisms, such as extreme dependence on the accuracy of camera poses and inability to effectively control erroneous Gaussian primitives densification caused by motion blur.
To solve these problems, we introduce a novel framework, Bi-Stage 3D Gaussian Splatting, to accurately reconstruct 3D scenes from motion-blurred images.
BSGS contains two stages. 
First, Camera Pose Refinement roughly optimizes camera poses to reduce motion-induced distortions. 
Second, with fixed rough camera poses, Global Rigid Transformation further corrects motion-induced blur distortions.
To alleviate multi-subframe gradient conflicts, we propose a subframe gradient aggregation strategy to optimize both stages.
Furthermore, a space-time bi-stage optimization strategy is introduced to dynamically adjust primitive densification thresholds and prevent premature noisy Gaussian generation in blurred regions.
Comprehensive  experiments verify the effectiveness of our proposed deblurring method and show its superiority over the state of the arts.

\end{abstract}



\maketitle


\section{INTRODUCTION}
\label{sec:INTRODUCTION}

Reconstructing a 3D scene from 2D images stands as a core issue in computer vision. 
Among various pioneering approaches, Neural Radiance Fields (NeRF) \cite{mildenhall2021nerf} has emerged as a prominent solution, demonstrating remarkable performance in recovering high-fidelity 3D scenes through its innovative differentiable volume rendering. 
Meanwhile, 3D Gaussian Splatting (3DGS) \cite{kerbl3Dgaussians} represents 3D scenes through Gaussian points and enables real-time rendering through efficient rasterization while preserving visual quality. 
However, both NeRF-based and 3DGS-based methods heavily depend on high-quality input images. In real-world scenarios, captured images are often affected by camera motion blur, which inevitably degrades reconstruction quality.
Camera motion blur is a prevalent form of image degradation, typically arising from camera jitter during exposure. Images affected by camera motion blur exhibit significant geometric inconsistencies between multi-view frames. Compounding this problem, the camera poses recovered from motion-blurred images using COLMAP \cite{schonberger2016structure} are often significantly biased, causing geometric misalignment when utilizing NeRF \cite{mildenhall2021nerf} and 3DGS\cite{kerbl3Dgaussians} for reconstruction. Furthermore, the sparse point clouds reconstructed from blurred images tend to contain substantial noise, severely affecting the quality of 3DGS initialization and resulting in marked degradation of performance.

Recently, several nerf-based approaches ~\cite{lee2024smurf, lee2023exblurf, wang2023badnerf, lee2023dp,peng2022pdrf} have emerged to process motion-blurred images to accurately reconstruct 3D scenes. 
Deblur-NeRF ~\cite{ma2022deblur} and DP-NeRF ~\cite{lee2023dp} attempt to model blur kernels and integrate physical priors, suffering from computationally intensive optimization processes. 
BAD-NeRF ~\cite{wang2023badnerf} and ExBluRF ~\cite{lee2023exblurf} jointly optimize camera poses and NeRF parameters, yet constrained by NeRF's inherent limitations, including low training efficiency.
3DGS-based methods ~\cite{chen2024deblur,zhao2024badgaussians,lee2024crim,oh2024deblurgs,lee2024deblurring,peng2024bags} have also incorporated motion blur modeling into their frameworks. BAD-GS \cite{zhao2024badgaussians} uses spline functions to represent camera trajectories during exposure time. DeBlur-GS \cite{chen2024deblur} derives camera pose derivatives to simplify the estimation of camera poses. 
However, existing methods often overlook that motion blur-induced errors in sparse point cloud initialization frequently cause premature and erroneous Gaussian splitting during training. 
In addition, their exclusive focus on camera pose optimization leads to excessive computational demands and heightened overfitting risks.
More critically, these methods are extremely sensitive to pose estimation accuracy where even minimal deviations can lead to severe degradation in reconstruction quality.

To address these challenges, we propose Bi-Stage 3D Gaussian Splatting (BSGS), a framework for robust 3D reconstruction from motion-blurred images.
BSGS includes two main steps: Camera Pose Refinement, which estimates approximate camera poses, and Global Rigid Transformation, where we apply global transformations to Gaussian points using the refined poses. In these two stages, we use weighted blending on subframes to mitigate gradient conflicts.

BSGS can effectively correct motion-induced distortions and reduce sensitivity to precise pose estimation.
For scene optimization, we minimize the photometric error between input blurred images and virtual synthesized images, supplemented by a max-pooling-based subframe gradient aggregation strategy to resolve multi-frame gradient conflicts. 
To further mitigate sparse point cloud initialization errors, we introduce Space-Time Coupled Densification. This adaptive thresholding mechanism adjusts Gaussian densification parameters dynamically, preventing premature artifact generation in blurred regions, outperforming fixed-threshold methods which previously methods used.

Our experimental results show that our BSGS effectively handles motion-blurred images, enabling the synthesis of sharp results in novel views. In summary, our main contributions are three-fold:
\begin{itemize}
    \item We introduce a two-stage training framework: Camera Pose Refinement to estimate camera poses; Global Rigid Transformation to simulate camera blur generation and correct motion-induced blur distortions.
    \item We develop a subframe gradient aggregation strategy to alleviate the gradient direction conflicts.
    \item We propose Space-Time Coupling Densification to prevent noise propagation and optimize Gaussian distribution.
\end{itemize}

\begin{figure*}[!t]
    \centering
    \includegraphics[width=0.995\linewidth]{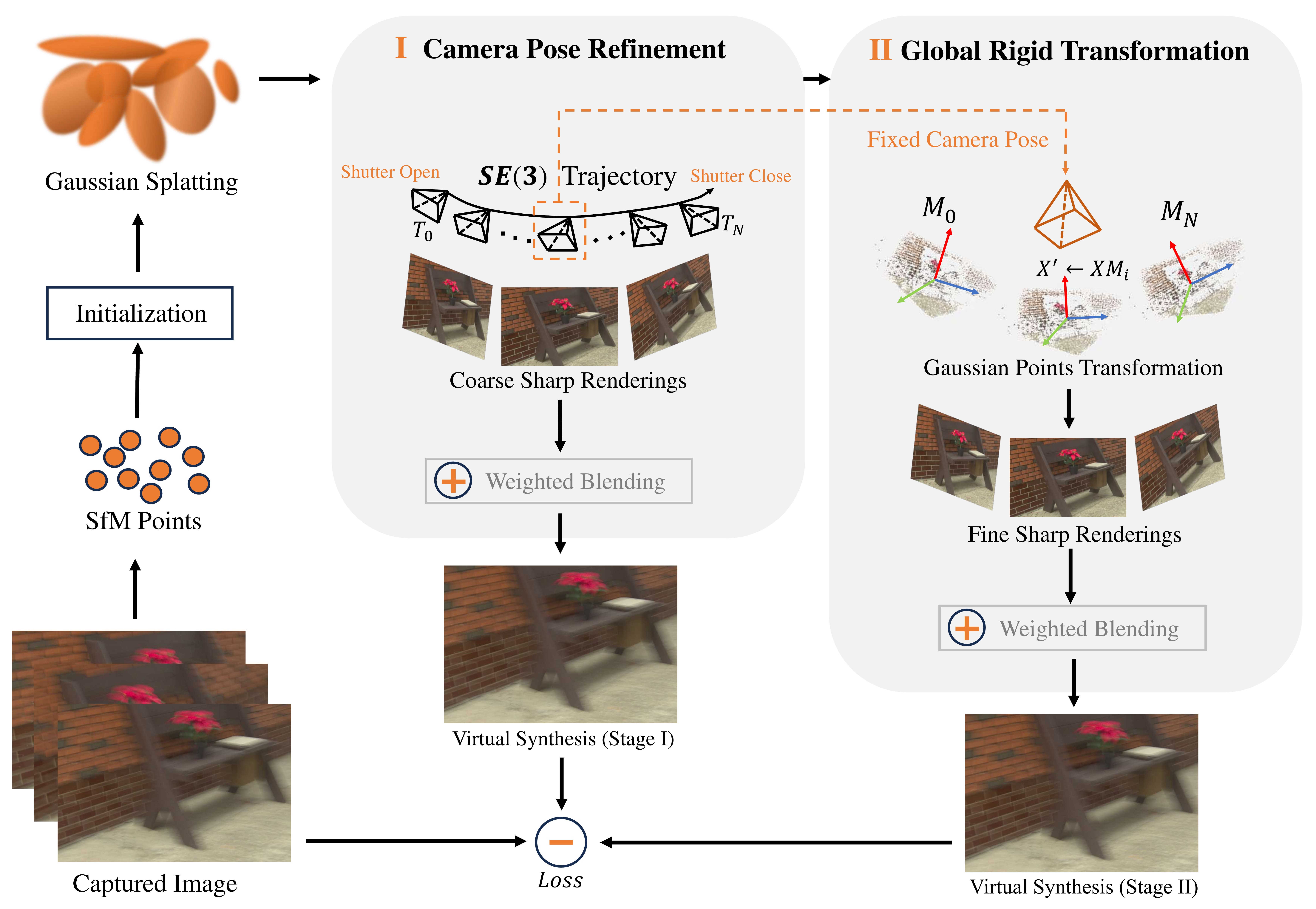}    
    \caption{The pipeline of our method.
We propose a Bi-stage approach for robust 3D reconstruction from motion-blurred images. In the first stage, we refine the camera poses to obtain an approximate estimate. In the second stage, we fix the optimized camera poses and apply global rigid transformations to the Gaussian points, simulating the motion blur. We optimize the Gaussian scene by minimizing photometric errors and introduce a subframe gradient aggregation strategy to resolve conflicts.
    }
    \vspace{-2 mm}
    \label{fig:pipeline}
\end{figure*}

\section{RELATED WORK}
\label{sec:RELATED WORK}

\subsection{Novel View Synthesis}
The emergence of Neural Radiance Fields (NeRF) \cite{mildenhall2021nerf} has significantly advanced the field of 3D vision, particularly in the context of novel view synthesis from multi-view images. 
However, in NeRF, the MLP layers required to evaluate the density and color at each pixel are computationally expensive, leading to slow rendering times. This has motivated further research into methods to accelerate both training and rendering processes ~\cite{chen2022tensorf,fridovich2022plenoxels,yu2021plenoctrees}. The advent of 3D Gaussian Splatting (3DGS) \cite{kerbl3Dgaussians} has significantly addressed this issue by introducing a explicit point-based representation and a novel differentiable splatting rasterizer, enabling real-time rendering without compromising visual quality. Subsequent works have extended NeRF and 3DGS to various domains, including dynamic 3D scene reconstruction ~\cite{li2022neural,li2021neural,pumarola2021d,park2021nerfies,park2021hypernerf,tretschk2021non}, human and facial reconstruction ~\cite{jiang2022neuman,gafni2021dynamic,peng2021animatable,weng2022humannerf}, and 3D content generation ~\cite{xu2024comp4d,zhou2024gala3d,EnVision2023luciddreamer,chung2023luciddreamer,tang2023dreamgaussian,yi2023gaussiandreamer,ouyang2023text,ling2024alignyourgaussians,yin20234dgen,ren2023dreamgaussian4d}. 

Despite the impressive advancements achieved by NeRF and 3DGS, most of these methods still rely on accurate camera parameters and sharp images, and struggle to effectively handle blurry input images. Our method addresses this challenge by incorporating camera motion blur deblurring into the 3D Gaussian Splatting framework, leveraging explicit point-based representations to recover sharp, high-quality scenes from blurry inputs while enabling real-time rendering of the reconstructed scenes.

\subsection{2D Image Deblurring}
Image deblurring is a fundamental problem in the field of image restoration, primarily addressing degradation caused by camera motion, defocus, or object movement during image acquisition.


Deep learning-based CNN methods ~\cite{kupyn2019deblurgan, nah2017deep, su2017deep, tao2018scale} learn end-to-end mappings but require large datasets and generalize poorly. Multi-scale approaches ~\cite{tao2018scale, gao2019dynamic, zamir2021multi} improve detail recovery but ignore 3D geometry, harming multi-view consistency. 

In contrast to the aforementioned 2D deblurring methods, our approach does not rely on a pre-trained network trained on large-scale datasets. Instead, we take advantage of multi-view consistency and employ a differentiable splatting rasterizer with a 3D Gaussian representation. This allows for high-quality 3D scene reconstruction and significantly improves the accuracy and robustness of the deblurring process.

\subsection{Radiance Field Deblurring}

Recently, deblurring methods ~\cite{chen2024deblur,ma2022deblur,peng2024bags,wang2023badnerf,lee2023exblurf,lee2023dp,oh2024deblurgs} based on NeRF and 3DGS have made significant progress in deblurring tasks. Deblur-NeRF ~\cite{ma2022deblur} embeds blind deblurring into NeRF, while DP-NeRF ~\cite{lee2023dp} models rigid blur kernels via 3D camera motion. BAD-NeRF \cite{wang2023badnerf}, BAD-GS \cite{zhao2024badgaussians} and Deblur-GS \cite{oh2024deblurgs} adopt physical modeling of camera motion and interpolate between predicted initial and final camera poses to optimize camera poses and scene representation parameters. Lee et al. \cite{lee2024deblurring} attributed the blur to the covariances of Gaussian points and proposed an MLP-based approach that predicts offsets for rotation and scale, effectively alleviating pixel-level blur effects. However, their method is limited to addressing defocus blur and does not account for motion-related blur.

\section{METHOD}
Given a sequence of motion-blurred images, our goal is to recover a sharp 3D scene representation. 
As shown in Fig. ~\ref{fig:pipeline}, our approach employs a bi-stage framework (BSGS) to deal with motion-blurred images, including Camera Pose Refinement and Global Rigid Transformation.
While the former optimizes the camera poses to achieve an approximate estimate, the latter then applies global rigid transformations to the Gaussian points and simulates the camera motion blur generation with fixed camera poses.
Furthermore, we propose the gradient aggregation strategy and the space-time coupling densification strategy to improve our scene reconstruction quality.
The details of each content will be presented in the following sections.

\subsection{Motion-Blurred Image Formation}
Motion blur in digital imaging arises from camera movement during the exposure period.  
The motion-blurred image $\mathbf{B}$ is expressed as an integral of sharp images across the exposure duration $\tau$:
\begin{equation}
    \mathbf{\hat{B}} = \phi \int_0^\tau \mathbf{C}_t \, dt, \label{eq:4}
\end{equation}
where $\phi$ is a normalization factor, and $\mathbf{C}_t$ represents the virtual latent sharp image captured at time $t \in [0, \tau]$ within the exposure. 
Eq. ~\ref{eq:4} captures the cumulative effect of light integration by the camera's photosensitive elements as the camera moves during exposure.

The extent of motion blur is intrinsically related to both the camera's velocity and the duration of exposure. 
Fast camera movements or short exposure usually result in an extremely low degree of blurring, while slow movements or long exposure times will increase the degree of blurring due to the increased relative motion amplitude.
In practice, the motion-blurred image can be computed as a weighted sum of sharp images sampled at discrete timestamps:
\begin{equation}
    \mathbf{\hat{B}}\approx \sum_{i=0}^{n-1} w_i \cdot \mathbf{C}_i. \label{eq:5}
\end{equation}
Here, $w_i$ denotes the weight associated with each sharp image $\mathbf{C}_i$ acquired at discrete time $t_i$. Our approach sets $w_i$ as a learnable parameter for optimizing the compositing effect of virtual sharp images. 
This physical model is fundamental for our work.

\subsection{Bi-Stage Deblurring Framework}
We parameterize the camera pose as \( T \in \mathbf{SE}(3) \) and employ three distinct interpolation schemes (linear, cubic spline, and B\'ezier) to model the camera's motion trajectory during the exposure interval. 
Given the initial camera pose \( T_{\text{start}} \) and the terminal pose \( T_{\text{end}} \) within the exposure period, the virtual camera pose at any intermediate timestamp \( t \in [0, \tau] \) can be formulated as:
\begin{equation}
T_t = T_{\text{start}} \cdot \exp\left( \frac{t}{\tau} \log\left(T_{\text{start}}^{-1} \cdot T_{\text{end}}\right) \right). \label{eq:6}
\end{equation}

As shown in Fig. ~\ref{fig:pipeline}, given \( N \) motion-blurred images $ \{\mathbf{B}_j\}_{j=1}^N $ and their corresponding inaccurate camera poses, we estimate a dedicated camera trajectory for each image \( B_j \) to model its specific motion characteristics during exposure. 
Following Eq. \ref{eq:5}, we render a sequence of images along the estimated camera trajectory through rasterization, which are then fused into a virtual synthesized image $\hat{B}$ via weighted blending with learnable parameters. 
Our 3D scene is optimized by minimizing the loss function as: 
\begin{equation}
\mathcal{L} = (1 - \lambda) \mathcal{L}_1 + \lambda \mathcal{L}_{D-\text{SSIM}}.
\label{eq:7}
\end{equation}

\subsubsection{\textbf{Camera Pose Refinement}}

To optimize the Gaussian parameters \( \theta \) and the camera poses \( T \) (i.e., \( T_{\text{start}} \) and \( T_{\text{end}} \)) for each image, we compute the gradient of the loss function \( \mathcal{L} \) with respect to the camera poses. This gradient is expressed as:
\begin{equation}
\frac{\partial \mathcal{L}}{\partial \mathbf{T}} = \sum_{j=0}^{N-1} \frac{\partial \mathcal{L}}{\partial \mathbf{\hat{B}}_j} \cdot \frac{1}{n} \sum_{i=0}^{n-1} \frac{\partial \mathbf{\hat{B}}_j}{\partial \mathbf{C}_i} \left( \frac{\partial \mathbf{C}_i}{\partial \mathbf{c}_i} \cdot \frac{\partial \mathbf{c}_i}{\partial \mathbf{T}} + \frac{\partial \mathbf{C}_i}{\partial \alpha_i} \cdot \frac{\partial \alpha_i}{\partial \mathbf{T}} \right),
\end{equation}

\begin{equation}
 \frac{\partial \alpha_i}{\partial \mathbf{T}} = \frac{\partial \alpha_i}{\partial \mathbf{o}_i} \cdot \frac{\partial \mathbf{o}_i}{\partial \mathbf{T}} + \frac{\partial \alpha_i}{\partial \mathbf{\Sigma}'_i} \cdot \frac{\partial \mathbf{\Sigma}'_i}{\partial \mathbf{T}} + \frac{\partial \alpha_i}{\partial \mu'_i} \cdot \frac{\partial \mu'_i}{\partial \mathbf{T}}
\end{equation}
where $\mathbf{\hat{B}}_j$ represents the virtual synthesized blurred images and \( \mathbf{C}_i \) the corresponding  rendered pixel color. 

Because the color term \( \mathbf{c}_i \) and the opacity term \( \mathbf{o}_i \) are independent of the camera pose, the two parameters can thus be ignored in the optimization process.
Due to superabundant matrix derivatives and transposition, the computational cost of calculating \( \frac{\partial \mathbf{\Sigma}'_i}{\partial \mathbf{T}} \) is prohibitively high ~\cite{yan2024gs}.
To alleviate it, we ignore covariance when optimizing camera poses.
Because the Gaussian enter has a much larger impact on the camera pose optimization than the covariance, ignoring the covariance has negligible impact on the camera pose optimization.
Finally, our gradient formula is expressed as:
\begin{equation}
\begin{aligned}
\frac{\partial \mathcal{L}}{\partial \mathbf{T}} 
= \sum_{j=0}^{N-1} \frac{\partial \mathcal{L}}{\partial \mathbf{\hat{B}}_j} \cdot \frac{1}{n} \sum_{i=0}^{n-1} \frac{\partial \mathbf{\hat{B}}_j }{\partial \mathbf{C}_i} \cdot \frac{\partial \mathbf{C}_i}{\partial \alpha_i}\cdot \frac{\partial \alpha_i}{\partial \mu_i} \cdot \frac{\partial \mu_i}{\partial \mu'_i} \cdot \frac{\partial \mu'_i}{\partial \mathbf{T}}.
\label{eq:l_T}
\end{aligned}
\end{equation}
where $\mu’_i = \mathbf{R}\mu_i + \mathbf{t}$, representing the transformation of the $i$-th Gaussian's center $\mu_i$ from world coordinates to camera coordinates, with $\mathbf{R} \in SO(3)$ denoting the rotation matrix and $\mathbf{t} \in \mathbb{R}^3$ the translation vector of the camera pose $\mathbf{T}$. During optimization, the camera pose $\mathbf{T}$ is iteratively adjusted to minimize the photometric discrepancy between virtual synthesized blurred images $\mathbf{\hat{B}}_j$ and the input blurry images $\mathbf{B}_j$, thereby refining the estimated camera trajectory.


\subsubsection{\textbf{Global Rigid Transformation}}

After the first stage of optimization, we obtain an approximately accurate camera pose corresponding to the midpoint of exposure time for each training image (referred to as BAD-GS \cite{zhao2024badgaussians}). We then fix these poses for further optimization. Based on Eq. \ref{eq:l_T}, we continue the derivation using:
\begin{equation}
\frac{\partial \mu'}{\partial \mathbf{T}} = \frac{\partial (\mathbf{K} \mathbf{T} \mu)}{\partial \mathbf{T} d},
\end{equation}
where \( \mathbf{K} \) represents the camera intrinsics matrix and \( d \) is the z-axis coordinate of the projection \( \mu \). Building upon the original camera pose, we introduce an additional global transformation matrix \( M_{trans} \in \mathbf{SE}(3) \) to control the global rigid transformation of the Gaussian points in the world coordinate. 

Finally, our bi-stage training mechanism is summarized as: 
\begin{equation}
\underbrace{\mathbf{K} (\mathbf{T}) \mu}_{\text{Stage I}} \xrightarrow{\text{freeze } \mathbf{T}} \underbrace{\mathbf{K} \mathbf{T} (M_{trans}) \mu}_{\text{Stage II}}.
\end{equation}
Stage I~(Camera Pose Refinement) only optimizes the camera poses $\mathbf{T}$ to achieve preliminary geometric alignment between consecutive input images.
Stage II~(Camera Pose Refinement) then freezes the approximately accurate camera poses and optimizes the global transformation $M_{trans}$ to further deblur 3D scenes.
Benefitting from the novel design, our method can significantly improve the accuracy and robustness of 3D reconstruction and enhance both computational efficiency and stability.



\subsubsection{\textbf{Subframe Gradient Aggregation via Max-Pooling}}
\label{sec:grad_aggregation}

\begin{figure}[t]
    \centering
    \includegraphics[width=1\linewidth]{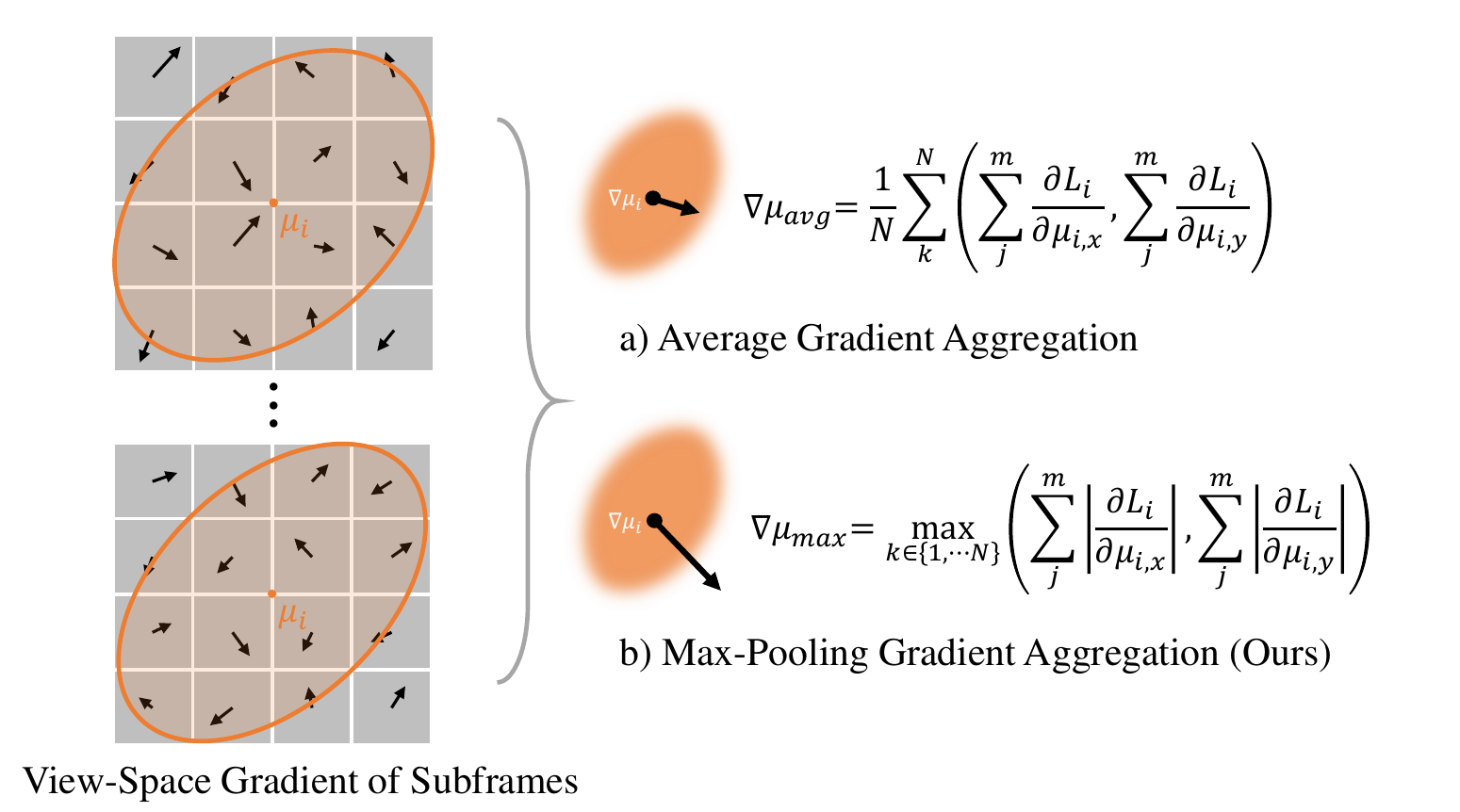}
    \caption{Subframe Gradient Aggregation via Max-Pooling.} 
    \label{fig:maxpool}
\end{figure}

To resolve gradient conflicts induced by motion blur and multi-view discrepancies, we propose a subframe gradient aggregation strategy that prioritizes dominant optimization directions. 

As shown in Fig. ~\ref{fig:maxpool}, during the bi-stage training process, for each input frame containing $n$ subframes, these subframes collectively constitute temporal slices of motion-blurred sequences, which are utilized to synthesize virtual motion-blurred images. We compute view-space positional gradients $\{\nabla{\mu}_1, \nabla{\mu}_2, \dots, \nabla{\mu}_n\}$ through differentiable rendering, where ${\mu} = (\mu_x, \mu_y)$ denotes the 2D Gaussian centers in view-space. 
These gradients are stacked to form aggregated gradient $\nabla{\mu}_{\text{agg}}$ via magnitude-maximizing pooling: 
\begin{equation}
    \nabla{\mu}_{\text{agg}}(x) = \max_{i \in \{1,\dots,n\}} \left|\nabla{\mu}_i(x)\right| \cdot \text{sign}\left(\nabla{\mu}_{i_{\text{max}}}(x)\right),
\end{equation}
where $x$ represents spatial coordinates and $i_{\text{max}}$ denotes the subframe index with maximum gradient magnitude at position $x$. 
This operation preserves the strongest gradient magnitude and prevents gradient cancellation while retaining directional consistency, effectively mitigating conflicting updates from different subframes.

\subsection{Space-Time Coupling Densification}

\begin{figure}[t]
    \centering
    \includegraphics[width=1\linewidth]{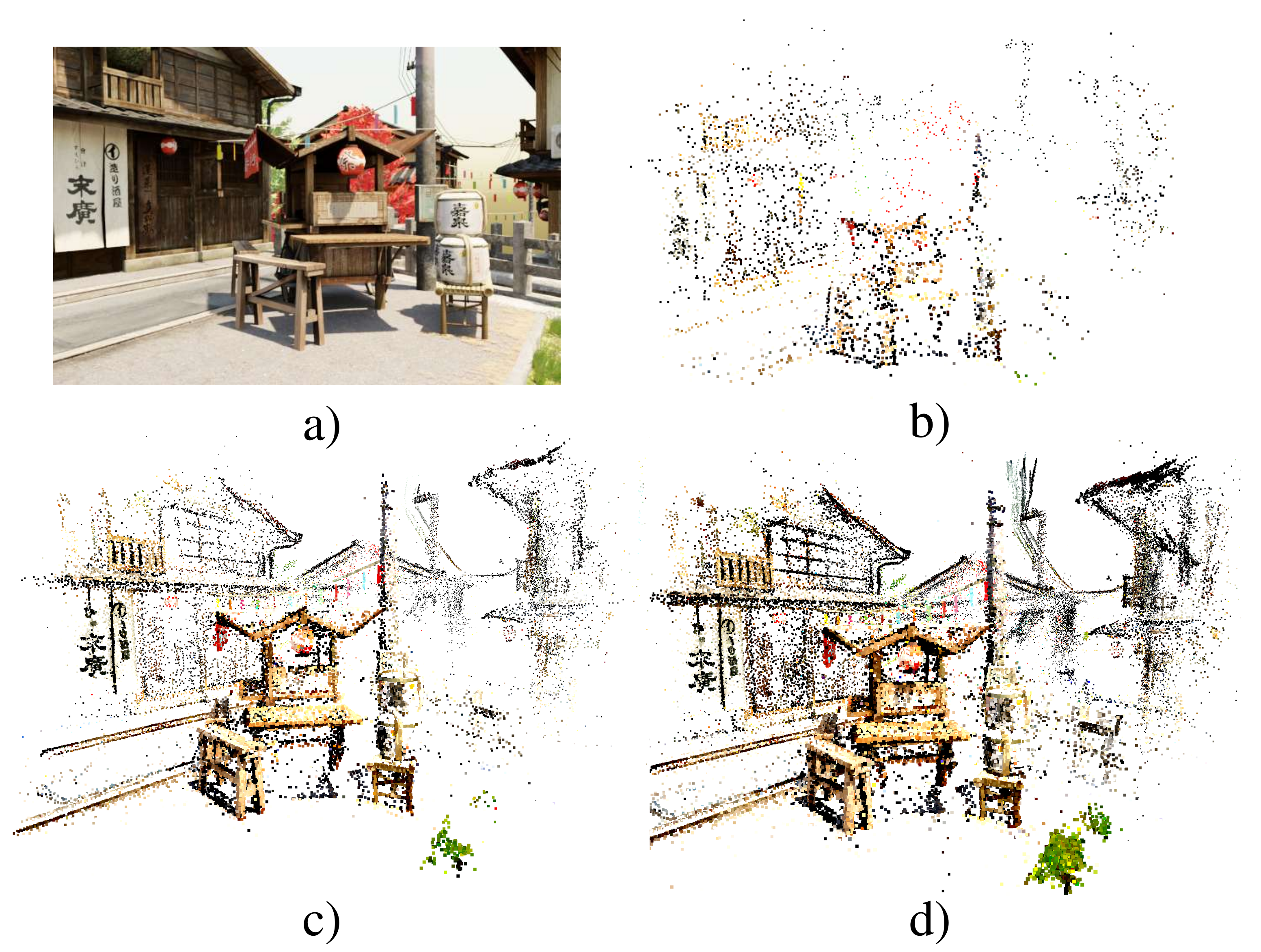}
\caption{Visualization of Space-Time Coupling Densification effects. a) The sharp 2D image to be reconstructed; b) The initialized point cloud; c) The point cloud obtained after applying the time-aware threshold $\tau_t$ and the space-aware threshold $\tau_s$ in Stage I, with the general scene outline and low-frequency foreground been roughly reconstructed; 
d) The final result after applying $\tau_t$ and $\tau_s$ in Stage II. The results demonstrate that Space-Time Coupling Densification can effectively control erroneous Gaussian primitives densification caused by motion blur and improve the reconstruction performance of detail information.}
    \label{fig:point_show}
\end{figure}

For 3DGS, signals of low frequency, such as smooth geometric structures, typically converge well during optimization, while signals of high frequency (e.g., details) often exhibit gradient oscillations. However, current 3DGS-based deblurring methods overlook important spatial variations among Gaussian primitives and progressive changes in blur levels throughout different training stages. These approaches only employ a fixed densification threshold for Gaussian splitting and cloning operations, resulting in unstable camera pose optimization due to their inability to adapt to the evolving reconstruction requirements.

To address these issues, we propose Space-Time Coupling Densification, a bi-stage dynamic coupling method designed to adaptively adjust the Gaussian densification threshold during training. 
In space level, because camera-motion blur tends to affect near-field Gaussian distributions with smaller depth and leads to unstable position estimation, we specially increase the densification threshold of these primitives to delay splitting. 
In contrast, Gaussian primitives with larger depth are not susceptible to camera-motion blur, we therefore lower their threshold to promote detail recovery.
In addition, different splitting times also affect the reconstruction performance.
With blurred images and inaccurate initial point cloud as input, premature splitting is prone to introduce errors.
To alleviate the problem, we choose to increase the densification threshold when refining the camera pose in Stage I. 
Once the camera pose is optimized, we lower the threshold when applying the global rigid transformation in Stage II to facilitate splitting and restore more high-quality details.

Finally, we formulate the space-aware threshold $\tau_s$ and the time-aware threshold $\tau_t$ as:

\begin{equation}
    \tau_s = \tau_0 \cdot \left(1 + \alpha \cdot e^{-\beta d}\right),
\end{equation}

\begin{equation}
    \tau_t(t) = 
    \begin{cases}
        \tau_0 \cdot (1 - \gamma t) & t \in Stage\  I \\
        \tau_0 \cdot \eta^{t-t_{split}} & t \in Stage\ II,
    \end{cases}
\end{equation}
where $\tau_0$ denotes the initial threshold, $\alpha$ controls the near-field gain, and $\beta$ regulates the depth decay rate. 
The space-aware threshold $\tau_s$ of near-field Gaussian primitives is significantly increased to suppress premature splitting, while $\tau_s$ of far-field primitives approaches the base value to promote detail recovery. 
The time-aware threshold $\tau_t(t)$ evolves throughout the whole training process, with $t$ denoting the current training step and $t_{split}$ marking the transition between stages. 
During Stage I, $\tau_t(t)$ undergoes linear decay modulated by $\gamma$, preventing excessive suppression. 
In Stage II, an exponential decay controlled by $\eta$ enables precise regulation of high-frequency components during Global Rigid Transformation.

Our final bi-stage dynamic densification threshold $\hat{\tau}$ is determined by combining the above two thresholds as:
\begin{equation}
    \hat{\tau} = \tau_s \cdot \tau_t(t)
\end{equation}
As shown in Fig. ~\ref{fig:point_show}, this joint optimization strategy effectively prevents the premature wrong generation of noisy Gaussians in blurred regions, leading to more high-quality reconstruction.


\section{EXPERIMENTS}
\begin{table*}[t]
  \centering
  \caption{Quantitative comparisons for deblurring on the real-world dataset of ExBluRF.}
  \label{tab:deblur_r}
  \resizebox{\textwidth}{!}{
    \begin{tabular}{l|ccc|ccc|ccc|ccc|ccc}
      \toprule
      & \multicolumn{3}{c|}{Bench} & \multicolumn{3}{c|}{Camellia} & \multicolumn{3}{c|}{Jars} & \multicolumn{3}{c|}{Jars2} & \multicolumn{3}{c}{Sunflowers} \\
      \cmidrule{2-16}
      & PSNR↑ & SSIM↑ & LPIPS↓ & PSNR↑ & SSIM↑ & LPIPS↓ & PSNR↑ & SSIM↑ & LPIPS↓ & PSNR↑ & SSIM↑ & LPIPS↓ & PSNR↑ & SSIM↑ & LPIPS↓ \\
      \midrule
      NeRF& 20.12 & 0.475 & 0.663 & 19.71 & 0.563 & 0.632 & 20.38 & 0.674 & 0.441 & 19.53 & 0.578 & 0.470 & 20.15 & 0.576 & 0.547 \\
      3DGS & 19.50 & 0.635 & 0.552 & 20.34 & 0.471 & 0.635 & 21.12 & 0.574 & 0.541 & 20.43 & 0.598 & 0.640 &  17.78 & 0.583 & 0.647 \\
      Deblur-NeRF & 26.15 & 0.721 & 0.241 &  27.33 & 0.713 & 0.319 & 27.12 & 0.703 & 0.235 & 26.64 & 0.729 & 0.298 &  30.10 & 0.814 & 0.172 \\
      BAD-NeRF & 24.93 & 0.642 & 0.363 & 25.19 & 0.644 & 0.298 & 27.95 & 0.741 & 0.257 & 25.56 & 0.721 & 0.267 & 27.69 & 0.670 & 0.213 \\
      ExbluRF & 27.44 & 0.732 & 0.215 & 24.79 & 0.702 & 0.237 &  25.16 & 0.678 & 0.214 & 26.63 & 0.698 & 0.319 & 27.62 & 0.805 & 0.232 \\
      Deblur-GS & 25.35& 0.684 &0.227  & 28.78 & 0.791 & 0.282  &  26.96 & 0.752 & 0.227 & 29.56 & 0.786 & 0.129 & 30.22 & 0.785 & 0.167 \\
      BAD-GS & 27.27 & 0.676 & 0.132 & 26.27 & 0.668 & 0.214 & 28.19 & 0.735 & 0.194 & 26.78 & 0.737 & 0.172 & 28.11 & 0.771 & 0.188 \\
      \cmidrule{1-16}
      \textbf{Ours} & \textbf{29.16} & \textbf{0.791} & \textbf{0.127} & \textbf{29.04} & \textbf{0.821} & \textbf{0.175} & \textbf{31.27} & \textbf{0.859} & \textbf{0.128} & \textbf{30.61} & \textbf{0.816} & \textbf{0.129} & \textbf{33.21} & \textbf{0.893} & \textbf{0.154} \\
      \bottomrule
    \end{tabular}
  }
\end{table*}

\begin{table*}[t]
  \centering
  \caption{Quantitative comparisons for deblurring on the synthetic dataset.}
  \label{tab:deblur_s}
  \resizebox{\textwidth}{!}{
    \begin{tabular}{l|ccc|ccc|ccc|ccc|ccc}
      \toprule
      & \multicolumn{3}{c|}{Cozyroom} & \multicolumn{3}{c|}{Factory} & \multicolumn{3}{c|}{Pool} & \multicolumn{3}{c|}{Tanabata} & \multicolumn{3}{c}{Trolley} \\
      \cmidrule{2-16}
      & PSNR↑ & SSIM↑ & LPIPS↓ & PSNR↑ & SSIM↑ & LPIPS↓ & PSNR↑ & SSIM↑ & LPIPS↓ & PSNR↑ & SSIM↑ & LPIPS↓ & PSNR↑ & SSIM↑ & LPIPS↓ \\
      \midrule
      NeRF&  23.13 & 0.652 & 0.255 & 22.71 & 0.546 & 0.469 & 27.12 & 0.742 & 0.275 & 20.53 & 0.477 & 0.403 & 21.55 & 0.764 & 0.472 \\
      3DGS & 24.50 & 0.630 & 0.285 & 21.29 & 0.557 & 0.412 & 28.62 & 0.768 & 0.253 & 19.37 & 0.450 & 0.395 & 20.73 & 0.624 & 0.388 \\
      Deblur-NeRF &  29.61 & 0.871 & 0.133 &  25.53 & 0.775 &0.267 & 30.92 & 0.843 & 0.135 & 23.58 & 0.729 & 0.298 & 26.34 & 0.814 & 0.177 \\
      BAD-NeRF & 30.77 & 0.876 & 0.092 & 28.69 & 0.832 & 0.183  & 32.39 &  0.840 & 0.117 & 25.33 & 0.804 & 0.175 & 26.79 &  0.841 & 0.085 \\
      ExbluRF &  28.56 &  0.891 &  0.142 & 26.92 & 0.814 & 0.231 &  27.30 & 0.782 & 0.243 & 26.62 & 0.855 & 0.220 & 24.63 & 0.762 & 0.232 \\
      Deblur-GS & 31.77 & 0.902 & 0.087 & 25.38 & 0.769 & 0.154 & 31.96 & 0.872 & 0.121 & 26.79 & 0.851 & 0.139 & 27.93 & 0.841 & 0.092 \\
      BAD-GS & 29.16 & 0.883 & 0.096 & 30.88 & 0.864 & 0.104 & 33.22 & \textbf{0.918} &\textbf{ 0.071} & 25.11 & 0.735 & 0.150 & 29.68 & 0.897 & \textbf{0.078} \\
      \cmidrule{1-16}
      \textbf{Ours} & \textbf{32.13} & \textbf{0.915} & \textbf{0.076} & \textbf{31.35} & \textbf{0.904} & \textbf{0.097} & \textbf{34.47} & 0.895 & 0.098 & \textbf{31.12} & \textbf{0.874} & \textbf{0.112} & \textbf{33.18} & \textbf{0.924} & 0.081 \\
      \bottomrule
    \end{tabular}
  }
\end{table*}

\begin{table*}[t]
  \centering
  \caption{Quantitative comparisons for novel view synthesis.}
  \label{tab:nvs_r}
  \resizebox{0.65\textwidth}{!}{
    \begin{tabular}{l|ccc|ccc|ccc}
      \toprule
      & \multicolumn{3}{c|}{Real Motion Blur} & \multicolumn{3}{c|}{Synthetic Motion Blur} & \multicolumn{3}{c}{Real-World Scene (ExBluRF)} \\
      \cmidrule(lr){2-4} \cmidrule(lr){5-7} \cmidrule(lr){8-10}
      & PSNR↑ & SSIM↑ & LPIPS↓ & PSNR↑ & SSIM↑ & LPIPS↓ & PSNR↑ & SSIM↑ & LPIPS↓ \\
      \midrule
      NeRF & 20.32 & 0.655 & 0.301 & 20.12 & 0.540 & 0.482 & 21.76 & 0.592 & 0.563 \\
      3DGS & 20.46 & 0.641 & 0.428 & 21.44 & 0.612 & 0.477 & 21.39 & 0.591 & 0.632 \\
      Deblur-NeRF & 26.57 & 0.744 & 0.179 & 23.88 & 0.754 & 0.288 & 28.87 & 0.709 & 0.402 \\
      BAD-NeRF & 25.26 & 0.745 & 0.215 & 28.32 & 0.823 & 0.382 & 25.31 & 0.647 & 0.376 \\
      ExBluRF & 23.41 & 0.675 & 0.219 & 27.81 & 0.834 & 0.232 & 26.94 & 0.659 & 0.472 \\
      Deblur-GS & 25.78 & 0.812 & 0.186 & 28.23 & 0.852 & \textbf{0.087} & 30.11 & 0.768 & 0.192 \\
      BAD-GS & 26.93 & 0.835 & \textbf{0.098} & 29.33 & 0.857 & 0.092 & 29.47 & 0.718 & 0.238 \\
      \cmidrule{1-10}
      \textbf{Ours} & \textbf{27.04} & \textbf{0.874} & 0.122 & \textbf{30.24} & \textbf{0.898} & 0.103 & \textbf{31.27} & \textbf{0.817} & \textbf{0.113} \\
      \bottomrule
    \end{tabular}
  }
\end{table*}
\subsection{Dataset}
We evaluate the performance of our method on real-world benchmark datasets, including Real-motion-blur \cite{ma2022deblur} and ExBluRF \cite{lee2023exblurf}, as well as a synthetic benchmark dataset derived from ExBluRF. The ExBluRF synthetic dataset, originally synthesized from Deblur-NeRF \cite{ma2022deblur}, was created using Blender \cite{blender} by averaging sharp virtual images captured during the exposure time, under the assumption of consistent camera motion velocity. The resolution for the multi-view images of ExBlur dataset is 800 × 540. Each scene in this dataset consists of 20 to 40 blurry images paired with corresponding sharp images. For the real-world benchmark datasets, the camera poses and sparse point clouds are obtained through Structure-from-Motion (SfM) using COLMAP \cite{schonberger2016structure}, with a set of training and testing images. Both the real-world and synthetic ExBluRF datasets present more challenging motion-blurred images, generated with randomly varying 6-DOF camera motion trajectories, and they contain more severe motion blur compared to the Real-motion-blur dataset. The public release of ExBluRF provides accurate camera poses and initial point clouds from the corresponding sharp image pairs using COLMAP. To validate the robustness of our method on real dataset, we perform COLMAP only on the blurry images in our experiments. This process results in inaccurate camera poses and sparse point clouds due to severe blurring, which are then used to initialize the 3DGS in subsequent experiments.

\subsection{Implementation Details}
We implemented our method based on the official implementation of the 3D-GS framework \cite{kerbl3Dgaussians} using PyTorch \cite{paszke2019pytorch}. The Gaussian scene parameters, camera pose parameters, and global transformation parameters are optimized using the Adam optimizer, with the learning rates for both the camera pose and global transformation optimizers exponentially decaying from \(1 \times 10^{-3}\) to \(1 \times 10^{-5}\). We set the number of virtual camera poses (i.e., \(n\) in Eq. \ref{eq:5}) in Stage I and the number of subframes in Stage II to 21 to balance performance and computational cost. For the synthetic dataset from ExBluRF \cite{ma2022deblur}, we employ linear interpolation in both stages. For the Real-motion-blur dataset \cite{ma2022deblur} and the real-world scenes from ExBluRF \cite{lee2023exblurf}, different interpolation methods are adopted depending on the specific scene characteristics. Initial estimates for the camera poses and Gaussian primitives are obtained using COLMAP \cite{schonberger2016structure}. All experiments were conducted on an NVIDIA RTX 3090 GPU.

\subsection{Experimental Results}

In our two-stage training framework, we first obtain a rough camera pose estimation by selecting the midpoint of the camera trajectory during the exposure period in Stage I. This optimized pose remains fixed throughout Stage II. 
To evaluate the effectiveness of our method, we select intermediate subframes from the sequence of high-quality sharp renderings generated through global rigid transformations, which are then aligned with the corresponding ground-truth images for quantitative and qualitative evaluation.
Specifically, we compare our BSGS with several related methods, including 3DGS \cite{kerbl3Dgaussians}, Deblur-NeRF \cite{ma2022deblur}, BAD-NeRF \cite{wang2023badnerf}, ExBluRF \cite{lee2023exblurf}, DeblurGS \cite{oh2024deblurgs}, and BAD-GS \cite{zhao2024badgaussians}.
Consistent with these approaches, we quantitatively assess the quality of our rendered image using PSNR, SSIM, and LPIPS. All metrics are computed by the API from TorchMetrics \cite{TorchMetrics}.

\subsubsection{\textbf{Quantitive And Qualitative Evaluation Results}}

Tab. \ref{tab:deblur_r} and \ref{tab:deblur_s} present quantitative comparisons for image deblurring, with the best results highlighted in bold. These tables demonstrate the comparative performance on both real-world datasets from ExBluRF and synthetic datasets from ExBluRF. It should be noted that the results of DeblurGS and BAD-GS were obtained using their officially released codes. Since the real-world datasets from ExBluRF contain severe motion blur, significant variations are observed in the experimental results. 

Tab. \ref{tab:nvs_r} provides quantitative comparisons for novel view synthesis on the Real-motion-blur datasets from Deblur-NeRF, as well as real-world and synthetic datasets from ExBluRF. Due to the absence of corresponding sharp ground truth images in the Real-motion-blur dataset, we only conduct novel view synthesis experiments. Compared with other 3DGS-based methods, our approach achieves superior performance in both deblurring and novel view synthesis.

\begin{figure*}[t]
    \centering
    \includegraphics[width=0.9\textwidth]{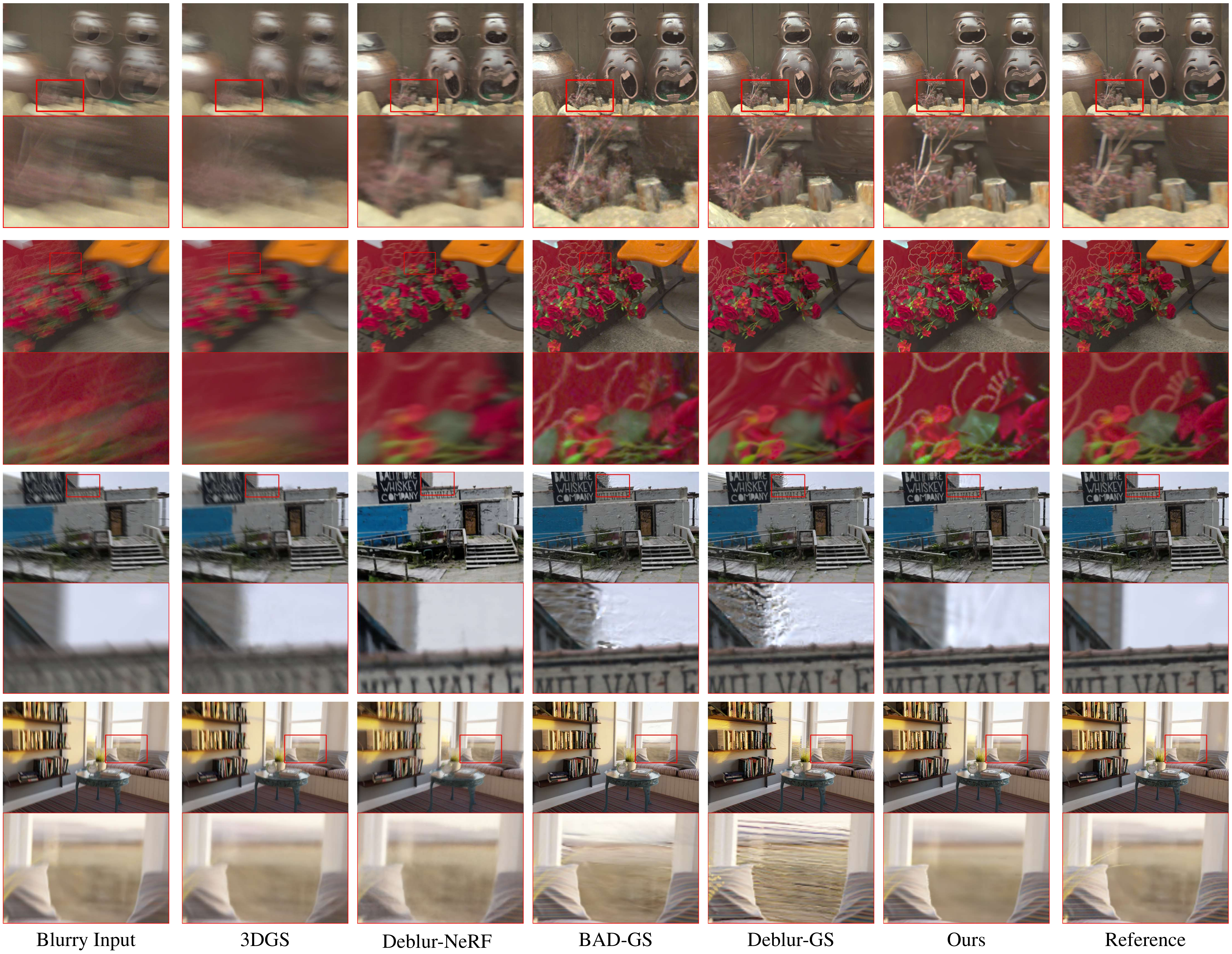}
    \caption{Qualitative comparison for deblurring on the synthetic and real-world scenes.
    The first two rows are from synthetic datasets while the last two rows are from real-world datasets. 
    The results demonstrate our method's superior deblurring and detail restoration performance compared to prior approaches.
    } 
    \label{fig:qualities_compare}
\end{figure*}

Fig. ~\ref{fig:qualities_compare} presents the qualitative comparisons between our method and other approaches. We selected four scenes from the ExBluRF dataset for evaluation, with camera poses of blurry input images initialized by COLMAP. Since BAD-NeRF relies on the linear trajectory to model camera motion, the method fails to restore sharp radiance fields when dealing with more severe hand-shake camera motion blur in real-world scenes. Experimental results demonstrate that our method is capable of reconstructing significantly sharper scenes. Further visualization results can be found in our supplementary materials.

\begin{figure}[t]
    \centering
    \includegraphics[width=0.95\linewidth]{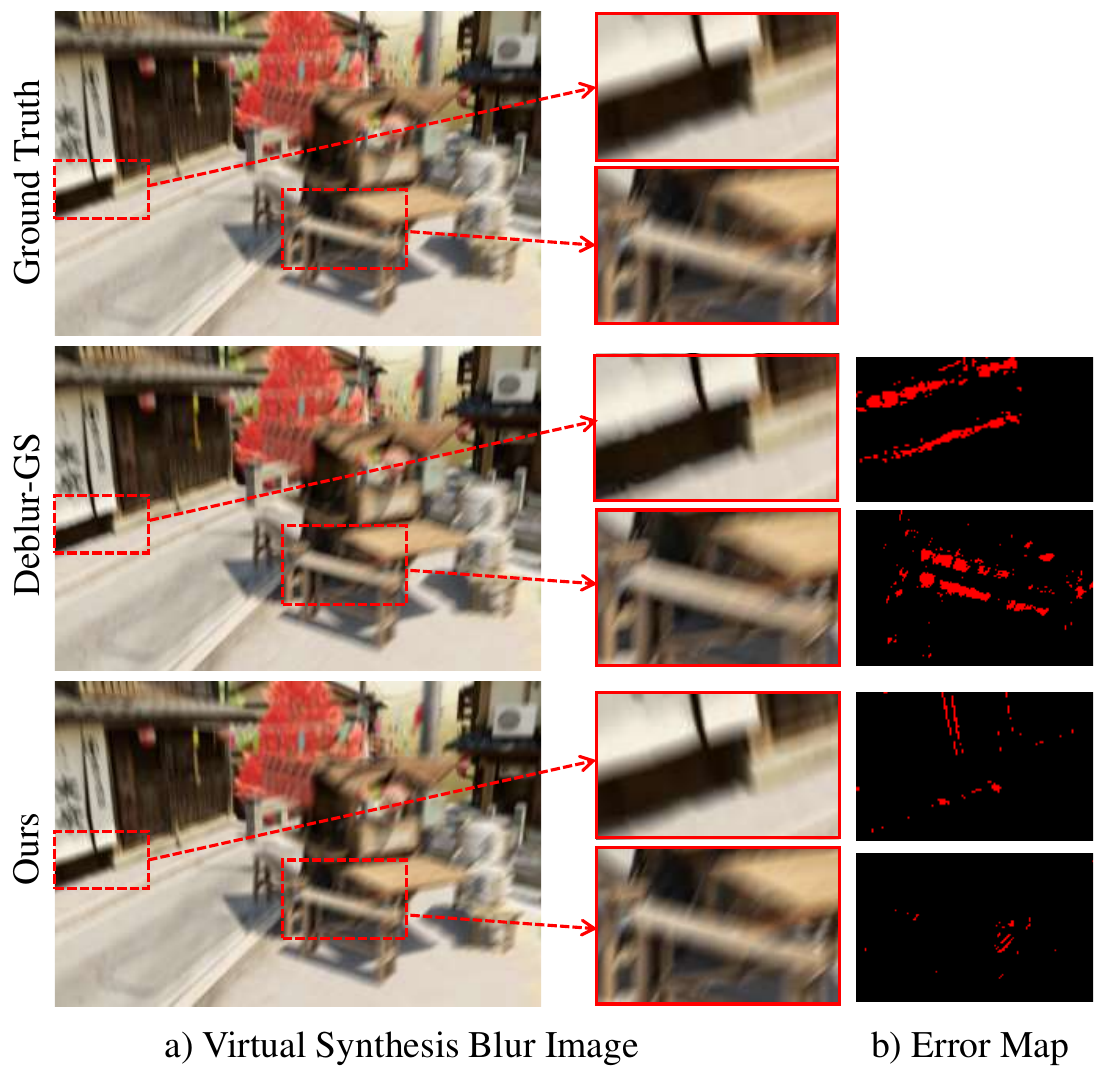}
    \caption{Visual comparison with error maps in \textbf{Trolley} scene.} 
    \label{fig:syn_compare}
\end{figure}

To further evaluate our method, we specially compute the error maps, as shown in Fig. ~\ref{fig:syn_compare}.
The error map is generated by computing the difference between the captured blurry image and the corresponding virtual image synthesized during the training process. The results show that our approach targets the finer details.

To evaluate the efficiency of our method in terms of computational performance and memory usage in 3D scene representation, we compare the Gaussian primitive numbers of different methods, as shown in Tab.~\ref{tab:point_number}.
The results show that our method significantly reduces the number of points while maintaining competitive deblurring performance. This reduction not only lowers memory consumption but also accelerates rendering and optimization processes, making our method more practical for real-world applications.

\begin{table}[t]
  \centering
  \caption{Comparison of primitive numbers for deblurring on the real and synthetic datasets of ExBluRF.}
  \label{tab:point_number}
  \resizebox{0.95\columnwidth}{!}{
    \begin{tabular}{c|cccc}
      \toprule
      & Camellia & Sunflowers & Tanabata & Trolley \\
      \midrule
      Deblur-GS  & 132406 & 154852 & 153094 & 164120 \\
      BAD-GS  & 225213 & 339439 & 539498 & 496786  \\
      \cmidrule{1-5}
      Ours  & \textbf{13446} & \textbf{40210} & \textbf{86442} & \textbf{100584} \\
      \bottomrule
    \end{tabular}
  }
\end{table}

In Tab.~\ref{tab:agg}, we compare the performance of different subframe gradient aggregation methods on on the real and synthetic dataset of ExBluRF.
"Average" represents the mean aggregation of all subframe gradients, whereas "Top-K" (where we set k=5 in our experiments) computes the average over the k largest gradients from the subframes.
 The results indicate that our max-pooling aggregation strategy shows superior performance across all evaluation metrics. This approach effectively preserves the most critical gradient features while robustly addressing gradient inconsistencies among subframes, thereby enabling more stable optimization and enhanced reconstruction quality for motion-blurred scenes.

\begin{table}[t]
  \centering
  \caption{Comparison of different subframe gradient aggregation methods for deblurring on the real and synthetic dataset of ExBluRF.}
  \label{tab:agg}
  \resizebox{0.45\textwidth}{!}{
    \begin{tabular}{c|ccc|ccc}
      \toprule
      \multirow{2}{*}{Method} & \multicolumn{3}{c|}{Stone\_Lantern} & \multicolumn{3}{c}{Tanabata} \\
      \cmidrule(lr){2-4} \cmidrule(lr){5-7}
      & PSNR$\uparrow$ & SSIM$\uparrow$ & LPIPS$\downarrow$ & PSNR$\uparrow$ & SSIM$\uparrow$ & LPIPS$\downarrow$ \\
      \midrule
      Average  & 28.50 & 0.769 & 0.181 & 30.04 & 0.819 & 0.152 \\
      Top-K (k=5) & 29.83 & 0.804 & 0.179 & 29.33 & 0.836 & 0.147 \\
      Max-Pooling  & \textbf{30.98} & \textbf{0.821} & \textbf{0.165} & \textbf{31.12} & \textbf{0.874} & \textbf{0.112} \\
      \bottomrule
    \end{tabular}
  }
\end{table}

\subsubsection{\textbf{Ablation Studies}}

We further evaluate the effectiveness of the two-stage approach and Space-Time Coupling Densification. As shown in Tab. \ref{tab:Ablation_method}, each component is fundamental for reconstructing sharp scenes, with the highest performance being achieved when all elements are integrated cohesively. Space-Time Coupling Densification contributes positively in both training stages.

\begin{table}[t]
  \centering
  \caption{Ablation studies on each element of our proposed method for deblurring on the real and synthetic dataset.}
  \label{tab:Ablation_method}
  \resizebox{\columnwidth}{!}{
    \begin{tabular}{ccc|ccc|ccc}
      \toprule
       \multicolumn{3}{c|}{Method} & \multicolumn{3}{c|}{Camellia} & \multicolumn{3}{c}{Tanabata} \\
      \cmidrule{4-9}
      Stage I & Stage II & STDC & PSNR↑ & SSIM↑ & LPIPS↓ & PSNR↑ & SSIM↑ & LPIPS↓ \\
      \midrule
      \checkmark &            &             & 26.59 & 0.712 & 0.225 & 28.39 & 0.683 & 0.276 \\
                 & \checkmark &             & 27.17 & 0.734 & 0.197 & 28.72 & 0.719 & 0.210 \\
      \checkmark &            & \checkmark  & 27.44 & 0.760 & 0.181 & 29.15 & 0.794 & 0.225 \\
                 & \checkmark & \checkmark  & 28.58 & 0.805 & 0.179 & 30.12 & 0.827 & 0.127 \\
      \checkmark & \checkmark &             & 27.70 & 0.751 & 0.183 & 28.94 & 0.725 & 0.169 \\
      \checkmark & \checkmark & \checkmark  & \textbf{29.04} & \textbf{0.821} & \textbf{0.175} & \textbf{31.12} & \textbf{0.874} & \textbf{0.112} \\
      \bottomrule
    \end{tabular}
  }
\end{table}

\begin{figure}[t]
    \centering
    \includegraphics[width=1.0\linewidth]{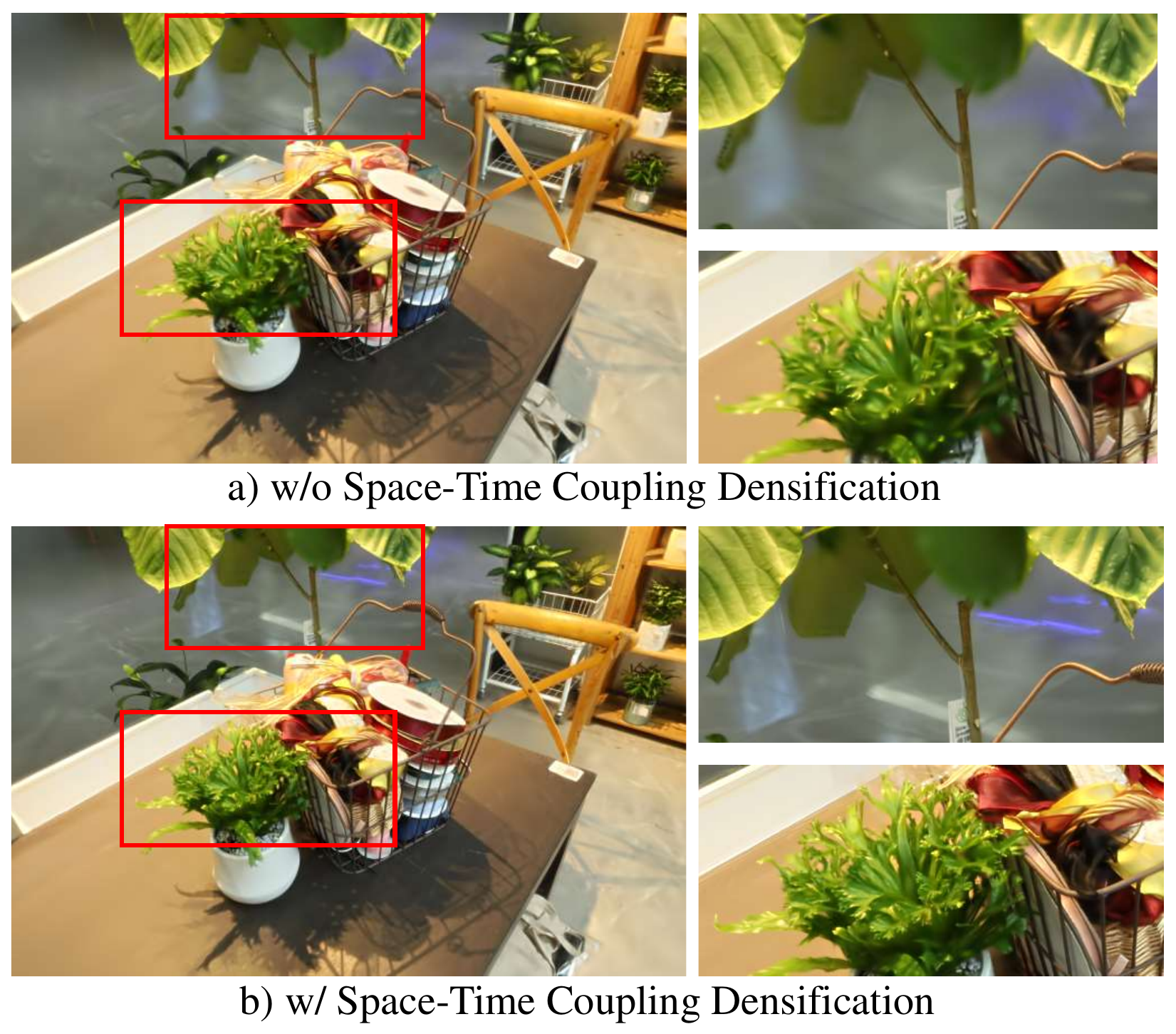}
    \caption{Ablation studies on Space-Time Coupling Densification.} 
    \label{fig:woST}
\end{figure}

Fig. ~\ref{fig:woST} compares the quality of reconstruction between using a constant threshold for Gaussian densification and those proposed in Space-Time Coupling Densification. 
It is clear that the rendering quality of the foreground remains comparable in both strategies.
However, our proposed Space-Time Coupling Densification stratery reconstructs more high-quality distant background regions, thus leading to a more detailed and coherent overall scene.

\section{Conclusion}
In this paper, we introduce a novel approach for camera motion deblurring in 3D Gaussian scene representation. 
Our approach employs a two-stage training trajectory to reconstruct sharper scenes with motion-blurred images. 
The first stage roughly optimizes the camera pose.
Then, with the fixed camera pose, the second stage applies global rigid transformations to the Gaussian primitives, simulating the process of camera motion blur generation.
Additionally, we introduce Space-Time Coupling Densification strategy to adaptively adjust the densification threshold, improving scene reconstruction quality. Experimental results demonstrate that our method outperforms all existing deblurring approaches, producing high-quality and robust scene representations.

\begin{acks}
This work was supported by the National Natural Science Foundation of China (No. T2322012, No. 62172218).
\end{acks}

\bibliographystyle{ACM-Reference-Format}
\balance
\bibliography{bib}


\clearpage

\appendix


\end{document}